\pdfoutput=1

\documentclass[11pt]{article}

\usepackage[final]{acl}

\usepackage{times}
\usepackage{latexsym}
\usepackage{amsmath,amssymb,amsthm}
\usepackage{graphicx}
\usepackage{ulem}
\usepackage{microtype}
\usepackage{hyperref}
\usepackage{url}
\usepackage{booktabs}
\usepackage{colortbl}
\usepackage{array}
\usepackage{lineno}
\usepackage{csquotes}
\usepackage{multirow}
\usepackage{amsfonts}
\usepackage{pifont}
\usepackage{enumitem}
\usepackage{tikz}
\usepackage{subfigure}
\usepackage{titletoc}
\usepackage{mathtools}
\usepackage{inconsolata}
\usepackage[ruled,vlined,linesnumbered]{algorithm2e}

\usepackage{tcolorbox}
\tcbset{
  colback=gray!5!white,
  colframe=gray!75!black,
  boxrule=0.5pt,
  arc=2pt,
  left=4pt,
  right=4pt,
  top=4pt,
  bottom=4pt
}

\usepackage[T1]{fontenc}

\usepackage[utf8]{inputenc}

\usepackage{microtype}

\usepackage{inconsolata}

\usepackage{graphicx}

\newcommand{\NC}[1]{\textcolor{red}{\textit{\textbf{[Cite]}}}}

\title{Mitigating Lost in Multi-turn Conversation via Curriculum RL with Verifiable Accuracy and Abstention Rewards}

\author{
  \textbf{Ming Li}\textsuperscript{1,2}\thanks{This work was completed during an internship at Amazon.},
  \textbf{Pei Chen}\textsuperscript{2},
  \textbf{Zhenhao Zhang}\textsuperscript{2},
  \textbf{Tao Yang}\textsuperscript{2},
  \textbf{Xinyang Zhang}\textsuperscript{2},
  \textbf{Han Li}\textsuperscript{2}, \\
  \textbf{Tianyu Cao}\textsuperscript{2}, 
  \textbf{Ming Zeng}\textsuperscript{2},
  \textbf{Zhuofeng Wu}\textsuperscript{2},
  \textbf{Meng Jiang}\textsuperscript{2},
  \textbf{Huasheng Li}\textsuperscript{2},
  \textbf{Lihong Li}\textsuperscript{2},
  \textbf{Bing Yin}\textsuperscript{2}\\
  \textsuperscript{1}University of Maryland ~~~~
  \textsuperscript{2}Amazon \\
  \texttt{minglii@umd.edu} ~~~~
  \texttt{ppeichen@amazon.com}
}

\begin{document}
\maketitle
\begin{abstract}
Large Language Models demonstrate strong capabilities in single-turn instruction following but suffer from Lost-in-Conversation (LiC), a degradation in performance as information is revealed progressively in multi-turn settings. Motivated by the current progress on Reinforcement Learning with Verifiable Rewards (RLVR), we propose Curriculum Reinforcement Learning with Verifiable Accuracy and Abstention Rewards (RLAAR), a framework that encourages models not only to generate correct answers, but also to judge the solvability of questions in the multi-turn conversation setting. Our approach employs a competence-gated curriculum that incrementally increases dialogue difficulty (in terms of instruction shards), stabilizing training while promoting reliability. Using multi-turn, on-policy rollouts and a mixed-reward system, RLAAR teaches models to balance problem-solving with informed abstention, reducing premature answering behaviors that cause LiC. Evaluated on LiC benchmarks, RLAAR significantly mitigates LiC performance decay (62.6\% to 75.1\%) and improves calibrated abstention rates (33.5\% to 73.4\%). Together, these results provide a practical recipe for building multi-turn reliable and trustworthy LLMs.
\end{abstract}

\section{Introduction}

\begin{figure}[t]
  \centering
  \includegraphics[width=1\linewidth]{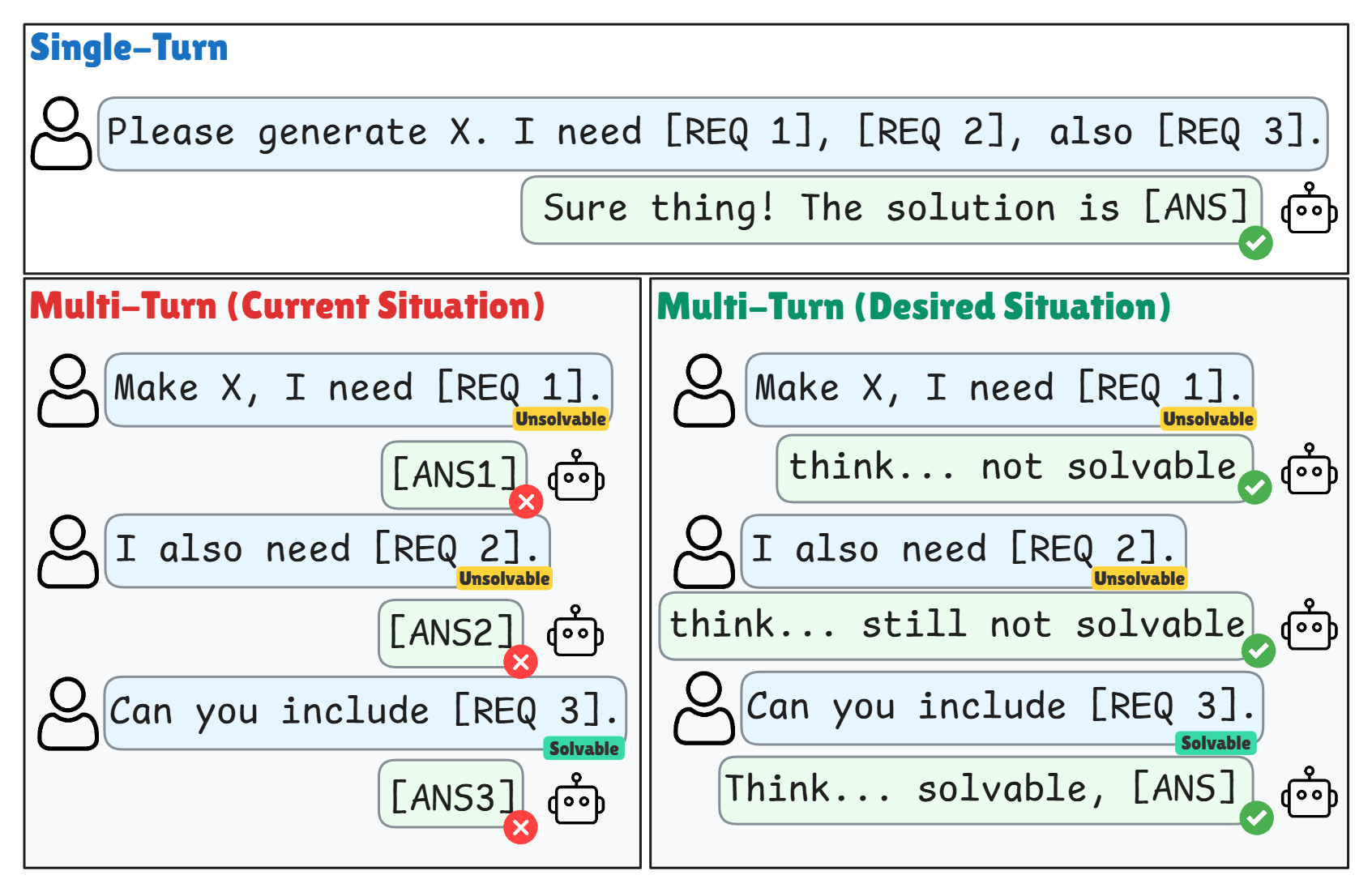}
  \caption{The common single-turn interactive setting with LLMs (upper) and the multi-turn setting proposed in ``Lost in Conversation'' (LiC). The multi-turn setting simulates the real-world scenario where the user gradually reveals the requirements of the task through multiple turns of dialogue with LLMs. \citet{laban2025llms} shows even the state-of-the-art models encounter severe performance declines. 
  The left panel shows the situation of most of the current LLMs, which try to solve the problem prematurely without checking the solvability of the question. The right panel shows the desired behavior of the model, where we novelly reveal the connections between LiC and models' abstention capability, which critically checks the solvability of the question and then solves the problem. } 
  \label{fig:intro}
  \vspace{-1em}
\end{figure}

Large Language Models (LLMs) have demonstrated a remarkable ability to comprehend and execute complex, fully-specified instructions in single-turn settings \cite{grattafiori2024llama3herdmodels, qwen2025qwen25technicalreport, guo2025deepseek, yang2025qwen3technicalreport}. This proficiency in instruction following has established them as powerful general-purpose problem solvers. However, this success covers a critical weakness that emerges in more natural, interactive scenarios: when instructions are revealed progressively in multi-turn conversations, models get ``Lost in Conversation'' (LiC) \citep{laban2025llms} (Figure \ref{fig:intro}). This phenomenon reveals a stark degradation in performance in multi-turn dialogues compared to single-turn baselines. The large-scale experiments show that this is not a failure of the models' underlying aptitude, as their best-case ability remains high, but rather a catastrophic increase in unreliability. In multi-turn settings, models frequently answer the question prematurely and subsequently fail to recover.

The LiC phenomenon stems from a set of identifiable behavioral flaws. Existing models lack the ability to abstain when facing unsolvable questions \cite{yin2023selfaware, fan2025missing,wen-etal-2025-know} and thus tend to generate full solutions prematurely, polluting the context with flawed assumptions and outputs (Figure \ref{fig:intro}, left). This behavior is exacerbated by failure to identify the crucial details from intermediate turns, extending the well-documented ``lost-in-the-middle'' \cite{liu-etal-2024-lost} problem from static long-context inputs to the temporal dimension of a dialogue. Solving this problem is critical, as gradual instruction refinement is the natural mode of interaction for complex, real-world tasks. The failure of LLMs to maintain reliability in these settings limits their adoption, serving as a major barrier for trustworthy AI \cite{li2022trustworthyaiprinciplespractices}.  

To overcome these behavioral patterns, a training paradigm is required to allow the model to explore conversational strategies rather than merely mimicking the static trajectories found in a supervised dataset. We therefore turn to Reinforcement Learning (RL) \citep{schulman2017proximal, ouyang2022training, guo2025deepseek}, but with a crucial adaptation for the multi-turn setting. Most existing RL applications treat prior turns as a fixed context for a single-step rollout, which still fails to capture the interactive dynamics of a dialogue. Our approach implements a \textbf{fully dynamic, multi-turn rollout} during policy optimization. In each training step, we simulate an entire conversation where the model's generated response at any given turn becomes part of the state for the subsequent turn. This design is critical: it allows the model to explore on every turn, and to learn from the downstream consequences of its intermediate actions. To ground this complex exploratory process, we use verifiable rewards (RLVR) \cite{guo2025deepseek}, such as correctness determined by a code interpreter or a math solver. These objective, ground-truth signals provide a stable and reliable learning target, which is essential for guiding the policy updates within the vast state space created by our dynamic rollouts.

However, an RL framework that optimizes solely for final task accuracy is itself insufficient. Such a setup implicitly encourages the model to attempt a full solution at every turn, as this is the only path to a potential reward, similar to the findings in \citet{kalai2025language}. This reinforces the very behavior of premature answering that causes LiC. The model pollutes the context with its own guesses, making it progressively harder to integrate the user's subsequent clarifications correctly. To counteract this, we introduce a crucial second component: an abstention reward. This novel reward signal creates an alternative, valuable action for the model: to recognize that the current information is insufficient and to deliberately output an abstention \citep{yin2023selfaware, fan2025missing}. By explicitly rewarding the model for knowing what it does not know, we train it to shift from a guess-and-check strategy to a more patient, information-gathering policy.

Finally, learning such a nuanced conversational policy, balancing task progression with judicious abstention, presents a significant credit assignment problem. In our multi-turn setting, a natural and effective proxy for this difficulty is the number of conversational turns. Training a model from scratch on long dialogues (e.g., five or more turns) is highly inefficient, as the probability of the model generating a correct, verifiable final response is exceedingly low. This results in a sparse reward landscape, making it difficult for the RL algorithm to learn an effective policy. We therefore introduce curriculum learning \cite{soviany2022curriculumlearningsurvey} to make this problem tractable. Our curriculum begins with short, simple conversations, where the connection between actions and a positive reward is more direct. This allows the model to first master the core skills of context integration and abstention in a high-signal environment. The conversational length and complexity are then progressively increased to medium and difficult scenarios, enabling the model to stably build upon its learned strategies. 

With the above discussions, we propose a novel framework, Curriculum \underline{\textbf{R}}einforcement \underline{\textbf{L}}earning with Verifiable \underline{\textbf{A}}ccuracy and \underline{\textbf{A}}bstention \underline{\textbf{R}}ewards (RLAAR), to mitigate the LiC problem. By comparing our method with baseline models and state-of-the-art models, we show that our method can significantly improve the performance and help the model to judge the solvability of the question in LiC settings.

The primary contributions of this work are:
\vspace{-0.5em}
\begin{enumerate}
    \item We propose a novel reinforcement learning framework that optimizes a model's conversational policy to mitigate LiC, moving beyond static imitation learning to foster more robust, exploratory decision-making.
    \vspace{-0.5em}
    \item We novelly \textbf{reveal the connection between LiC and models' abstention capability} and design a \textbf{verifiable, mixed-reward system} that uniquely combines accuracy rewards for task completion with abstention rewards for handling ambiguity.
    \vspace{-0.5em}
    \item We demonstrate that a \textbf{curriculum learning strategy} is critical for training stability, enabling the model to learn complex multi-turn behaviors by progressively increasing conversational depth and difficulty.
\end{enumerate}

\section{Related Works}

\subsection{Multi-turn Conversation}

As interest in multi-turn evaluation surged alongside the meteoric rise of ChatGPT, early widely used efforts like MT-bench~\cite{zheng2023judging} relied on crowd-sourced annotations to assess LLM-as-a-judge performance. Subsequent work extended MT-bench by supporting longer dialogues~\cite{kwan2024mt,duan2024botchat}, offering finer-grained assessment~\cite{bai2024mt}, and targeting new axes such as naturalness~\cite{sirdeshmukh2025multichallenge} or tool use~\cite{2024bfcl,wang2024mint}. While these tasks require some context tracking, they rarely demand the active integration needed to resolve underspecified user instructions. Such underspecification is both common in real human-AI exchanges~\cite{herlihy2024overcoming} and a natural conversational tendency captured by the ``principle of least effort''~\cite{zipf1949human}. Scaling multi-turn experimentation, therefore, requires simulating the user. Prior studies have done so via templates~\cite{choi2018quac,reddy2019coqa,laban2023you,deng2024multi}, LLMs~\cite{poelitz2025synthetic,li2024iqa,chang2025chatbench,liang2024mathchat}, human annotators~\cite{finch2023don,chang2025chatbench}, or real users in controlled studies~\cite{ram2018conversational,laban2021s,chiang2024chatbot}. While real users yield the most natural interactions, this approach sacrifices scalability and reproducibility. LiC therefore adopts an LLM-based simulator to balance control with diversity, enabling the study of \textit{LLM behavior} in multi-turn contexts and exposing the core issues underlying the LiC phenomenon, the central focus of this work.

\subsection{Response Abstention}

One key capability for mitigating the Lost in Conversation problem is an LLM's ability to judge a query's solvability and abstain from unsolvable ones, so the model can, to some extent, avoid issuing premature, verbose answers that further pollute the context. Numerous datasets evaluate abstention, but they typically target a single problem type, e.g., unanswerable questions \citep{yin2023selfaware,amayuelas2024kuq}, multiple-choice questions with no correct option \citep{madhusudhan2025llms}, or underspecification \citep{slobodkin2023musique-nq,zhang2024clamber,li2025questbench}. Closely related is verbalized uncertainty \citep{lin2022teaching,tian2023just}, the model's explicit expression of doubt as a downstream signal that it cannot appropriately answer. Several works report limited performance and generalization for verbalized uncertainty as an uncertainty-quantification method \citep{vashurin2024benchmarking, lin2022teaching, xiong2024can}. At the same time, \citet{kapoor2024large} shows that fine-tuning can improve verbalized uncertainty, and \citet{kadavath2022language} demonstrates that suitable prompting can elicit correctness probabilities that become increasingly calibrated as models scale. Parallel efforts have improved abstention via fine-tuning \citep{chen2024teaching,brahman2024art} and through explanation generation \citep{deng2024dontjustsay}. Prior work has also benchmarked and aimed to improve policy compliance \citep{brahman2024art, DBLP:conf/aies/MuellerGBPP24, mazeika2024harmbench}.
Recently, as the growing interest in Large Reasoning Models, MiP-Overthinking \citep{fan2025missing} further reveals an interesting phenomenon that a longer thinking process not only does not help the model to abstain from the unsolvable question, but also exacerbates the problem. In our work, our method has a better abstention ability in this LiC setting. 
\looseness-1

\subsection{RL in LLMs}

Reinforcement Learning has become a powerful paradigm for aligning LLMs with human preferences and values. The most widely used method is Proximal Policy Optimization (PPO)~\cite{ouyang2022training, schulman2017proximal}. Building on PPO, GRPO removes the need for an explicit value function by estimating advantages from group scores and has become a popular training paradigm~\cite{shao2024deepseekmath}. Reinforcement Learning with Verifiable Rewards (RLVR) further streamlines training by evaluating outputs with predefined rules, eliminating the reward model~\cite{guo2025deepseek}. Variants that address specific limitations of GRPO have also been proposed~\cite{su2025crossing}. To mitigate length bias, Dr. GRPO drops normalization terms, while SRPO adds two-stage training with historical sampling~\cite{liu2025understanding, zhang2025srpo}. To improve training stability, DAPO \cite{yu2503dapo} increases clip rates, and GVPO \cite{zhang2025srpo} provides an analytical solution to KL-constrained reward maximization. Compared with prior work, our method utilizes multi-turn rollouts and a mixed reward system to train the model to mitigate the LiC problem.

\section{Methodology}

\subsection{Problem Formulation}
We cast multi-turn, incrementally specified dialog tasks defined in \citet{laban2025llms} as a finite-horizon Markov Decision Process with a verifiable terminal signal. 
A complete task \(T\) is represented as an ordered sharded question sequence \(\mathcal{S}=\{s_1,\dots,s_K\}\), where each shard question \(s_k\) is a subset of the original complete question $q$ and encodes the user's newly revealed constraints at round \(k\). At turn \(k\), the policy \(\pi_\theta\) consumes the context \(c_k=(s_1,a_1,\ldots,s_{k})\) and emits a textual action \(a_k\). At the final round \(K\), the model returns a terminal answer \(a_K\).
The environment provides a single terminal reward, either from an accuracy verifier (solvable cases) or from an abstention verifier (unsolvable cases). 

For a trajectory \(\tau=(c_1,a_1,\ldots,c_K,a_K)\), we define
{\small
  \begin{equation}
   r(\tau)=
   \begin{cases}
   r_{\mathrm{acc}}(a_K;T)\in\{0,1\}, & \text{if } T \text{ is solvable}, \\
   r_{\mathrm{abs}}(a_K;T)\in\{0,1\}, & \text{if } T \text{ is unsolvable}. \\
   \end{cases}
  \end{equation}
  \label{eq:reward}
}
$r_{\mathrm{acc}}(a_K;T)$ represents the verifiable accuracy reward and $r_{\mathrm{abs}}(a_K;T)$ represents the verifiable abstention reward  for the terminal answer $a_K$ on the task $T$. 
The overall pseudo code of the RLAAR method is provided in Appendix \ref{sec:pseudocode}.

\subsection{Curriculum RL with Verifiable Accuracy and Abstention Rewards}

\subsubsection{Multi-turn Rollout}

Different from most of the existing RL methods for LLMs that only focus on the single-turn setting, our method novelly utilizes on-policy multi-turn rollouts that explicitly simulate the dialogue process. 
Specifically, there are three different types of rollouts during the training:
\begin{itemize}
  \item \textit{Solvable-Single}: the model is given the original full question and expected to generate the final answer in one action.
  \item \textit{Solvable-Multi}: the model is given one of the sharded questions each turn until all shards are revealed and expected to generate the final answer after multiple actions.
  \item \textit{Unsolvable-Multi}: the model is given one of the sharded questions each turn, while some shards are intentionally withheld and expected to \textit{abstain} from the unsolvable question.
\end{itemize}

\paragraph{\textit{Solvable-Single.}}
In this setting, the episode consists of a single turn ($K=1$). The complete task description $q$ is given as the first and only shard, $s_1=q$. The policy generates a single action $a_1 \sim \pi_\theta(a|s_1)$. The resulting trajectory is $\tau = (s_1, a_1)$, and the policy is optimized to maximize $\mathbb{E}[r_{\mathrm{acc}}(a_1; T)]$. This setting is equivalent to the single-turn setting in the existing RL methods for LLMs. It is only utilized at the beginning of the training. 

\paragraph{\textit{Solvable-Multi.}}
In this setting, the full sequence of $K$ shards, $\mathcal{S} = \{s_1, \ldots, s_K\}$, is revealed to the model sequentially. 
At the beginning of each turn $k \in \{1,\dots, K\}$, the environment presents the next shard $s_k$. The policy then conditions on the accumulated context $c_k=(s_1,a_1,\ldots,s_{k-1},s_k)$ to sample an action $a_k \sim \pi_\theta(a|c_k)$.
This process repeats until the full trajectory $\tau=(s_1, a_1, \dots, s_K, a_K)$ is generated. The optimization objective is to maximize the expected final reward $\mathbb{E}[r_{\mathrm{acc}}(a_K; T)]$. This setting is directly aligned with the evaluation setting presented in LiC, and thus serves as the main training setting for the model.

\paragraph{\textit{Unsolvable-Multi.}}
This setting is similar to the \textit{Solvable-Multi} case, but the interaction is truncated after $M < K$ turns. 
The environment provides shards $s_1, \ldots, s_M$ sequentially, and the policy generates corresponding actions $a_1, \ldots, a_M$. After the environment presents the final shard $s_M$ and the policy produces action $a_M$, the episode terminates. The resulting trajectory is $\tau=(s_1, a_1, \dots, s_M, a_M)$, and the policy is trained to produce a terminal action $a_M$ that explicitly abstains, maximizing the expected reward $\mathbb{E}[r_{\mathrm{abs}}(a_M; T)]$. Different from the above settings, which focus on the model's correctness, this setting is designed to focus on the model's capability to justify the solvability of the question. It also serves as the main training setting.

In the main training process, we further introduce the abstention ratio $m \in [0, 1]$ to control the proportion of the \textit{Unsolvable-Multi} rollouts. When $m=0$, the model is only trained on the \textit{Solvable-Multi} rollouts. When $m=1$, the model is only trained on the \textit{Unsolvable-Multi} rollouts. This hyperparameter is critical for the model to learn the balance between correctness and abstention. 
Note that for both the multi-turn settings, the rollout process is fully \textit{dynamic and on-policy}, rather than just using the static trajectory as the context. Thus, the model can explore on every turn and learn from the downstream consequences of its intermediate actions.

\subsubsection{Mixed Rewards}
A standard RLVR \citep{guo2025deepseek} setup that optimizes solely for question accuracy is insufficient for the LiC problem. In fact, such a reward scheme would implicitly encourage the model to attempt a full solution at every turn, as this is the only path to a potential reward, rather than to first evaluate the solvability of the current context. This reinforces the very behavior of premature answering we aim to eliminate. 
To address this, we introduce a mixed-reward system that provides distinct, verifiable signals for two contrary but equally desirable behaviors: question accuracy and strategic abstention. As defined in Equation~\ref{eq:reward}, our reward function relies on two verifier functions, $V_{\text{acc}}$ and $V_{\text{abs}}$, which formalize the reward calculation.

\paragraph{\textit{Accuracy Reward ($r_{\text{acc}}$)}.} 
This reward is the standard accuracy reward used for existing RLVR methods, which rewards the model for correctly solving the question.
In our setting, it is applied to both the \textit{Solvable-Single} and \textit{Solvable-Multi} settings.
Specifically, we define an automatic accuracy verifier, $V_{\text{acc}}(a_K, T)$, which returns `true' if the model's final answer $a_K$ correctly solves the complete task $T$, and `false' otherwise. The reward is then formally defined with an indicator $\mathbb{I}(\cdot)$:
\begin{equation}
r_{\text{acc}}(a_K; T) = \mathbb{I}(V_{\text{acc}}(a_K, T) = \text{true})
\end{equation}
This assigns a reward of 1 for a correct answer and 0 otherwise.
Note that for different types of tasks, the verifier $V_{\text{acc}}$ is different: For the math tasks, the verifier checks if the model's answer is the same as the ground truth answer and wrapped in the predefined format, i.e., ``\boxed{x}''. For the code tasks, the verifier checks if the model's generated code is executable and can successfully pass the test cases.

\paragraph{\textit{Abstention Reward ($r_{\text{abs}}$)}.} 
This reward is designed for the \textit{Unsolvable-Multi} setting, aiming to train the model to abstain from the unsolvable question. 
The reward is determined by a simple verifier, $V_{\text{abs}}(a_M)$, which checks if the model's terminal action $a_M$ contains the predefined string representing the abstention behavior, i.e., ``\boxed{Abstain}''. 
The reward is task-agnostic and only depends on the model's terminal action. The reward is then:
\begin{equation}
  r_{\text{abs}}(a_M; T) = \mathbb{I}(V_{\text{abs}}(a_M) = \text{true})
\end{equation}
This assigns a reward of 1 for correctly abstaining and 0 for any other action, such as attempting to provide a premature answer. Since only string-matching is required for this reward, the calculation is efficient. 

The combination of these two reward signals is crucial. It creates a balanced incentive structure that teaches the model a critical meta-skill: to first evaluate the solvability of the current context before committing to an answer. By providing a valuable alternative to guessing (i.e., abstaining for a reward), the model learns to handle the ambiguity inherent in multi-turn, incrementally specified dialogues, which is the key to mitigating the LiC phenomenon, which is important to trustworthy AI. 
\looseness-1

\subsubsection{Curriculum Design}

Training an RL policy on long multi-turn dialogues from scratch is naturally difficult. The probability of generating a correct final response after a long sequence of intermediate actions is exceedingly low, leading to a sparse reward landscape that makes the learning process unstable and inefficient. To address this challenge, we introduce a curriculum learning strategy that makes the complex learning problem tractable by gradually increasing the conversational difficulty.

Our curriculum structures the learning process by naturally defining difficulty based on the number of shards $K$ the original complete question is segmented into. 
A higher K corresponds to a longer and more challenging conversation, which is naturally more difficult to solve. The curriculum unfolds in three main stages.

\paragraph{\textit{Stage 1: Threshold Establishment.}} 
The training process begins exclusively with the \textit{Solvable-Single} setting. The model is trained on original, complete problems where $K=1$. After an initial training period, we calculate the moving-average reward $\bar{r_0}$ over a predefined window ($5$ by default). This value serves as a dynamic, model-specific performance baseline that reflects the model's initial performance on the easiest problems. With an additional threshold ratio $\rho \in [0,1]$, we use $\rho \times \bar{r_0}$ as the real threshold for the subsequent stages. 

\paragraph{\textit{Stage 2: Main Training.}} 
After establishing the baseline reward, the curriculum introduces multi-turn tasks, \textit{Solvable-Multi} and \textit{Unsolvable-Multi}, starting with $K=2$ for each training batch. 
The model continues to train at a given difficulty level $K$ until its current moving-average reward over a window surpasses the predefined threshold. 
Formally, the curriculum progresses from $K$ shards to $K+1$ shards only when the condition $\bar{r} \geq \rho \times \bar{r_0}$ is met. This process repeats until a predefined maximum shard count, $K_{\text{max}}$, is reached.
The reason that an additional threshold ratio $\rho$ is introduced is that the model's performance on harder scenarios is naturally lower than in the single-turn setting. Thus, if no such scaling is applied, the model may get stuck and never progress to the next difficulty level.

\paragraph{\textit{Stage 3: Randomized Training.}}
Once the model has successfully passed the maximum difficulty level, the curriculum transitions to its final stage. 
In this phase, for each training batch, the number of shards $K$ is randomly sampled between $1$ and $K_{\text{max}}$. This randomization encourages a more robust and generalized policy that can handle conversations of varying lengths, rather than overfitting to the maximum difficulty level.


\section{Experiments}

\subsection{Implementation Details}

\paragraph{Models.}
We implement our method on Qwen3-8B \cite{yang2025qwen3technicalreport}, Qwen3-1.7B, Qwen2.5-7B-Instruct \cite{qwen2025qwen25technicalreport}. All the models are instruction-aligned versions. During training, we use the default chat template for each model. For the Qwen3 model families, we utilize the non-reasoning version of the conversation template for both training and testing.

\paragraph{Data.}
Our data mainly includes two subcategories: the math dataset and the code dataset. For the Math dataset, we utilize the GSM8K \cite{cobbe2021trainingverifierssolvemath} dataset as our source dataset. For the Code dataset, we utilize the Eurus \cite{cui2025processreinforcementimplicitrewards} code subset, including code generation problems sourced from TACO \cite{li2023taco}, Codecontests \cite{li2022competition}, Codeforces, and apps \cite{hendrycks2021measuring} datasets. For each math or codesubcategory, we randomly sample 1000 problems that are long enough to be segmented into multiple sharded instructions for training.

\paragraph{Training Details.}
Our code is implemented based on the Verl \footnote{\url{https://github.com/volcengine/verl}} framework, and we conduct all the experiments on 8 A100 GPUs. 
We use the AdamW optimizer with a learning rate of 1e-4 and a weight decay of 0.1. We use a batch size of 64 for training. The experiments are conducted on the GRPO \cite{shao2024deepseekmath} algorithm by default. 
In GRPO, the group size is set to $4$, and we use the default settings for the other hyperparameters. 
The training utilizes a dynamic model-specific stopping strategy, where the maximum training step is set to $100$ steps after the model gets into the Randomized Training stage (stage 3). 
For method-specific parameters, the default maximum shard number is $K_{\text{max}}=5$; the default abstention ratio is $m=0.1$; the default threshold ratio is $\rho=0.8$. 

\paragraph{Evaluation.}
To evaluate the performance of our method, we focus on \textit{\textbf{LiC Score}} and the \textit{\textbf{Abstain Score}}. The \textit{LiC Score} is the ratio of the problems that can be correctly solved in the concatenated instruction setting but not in the sharded instruction setting, which is calculated as $\frac{Accuracy(Sharded)}{Accuracy(Concatenated)}$. This is the standard metric used in LiC and presented as the main findings that even the recent state-of-the-art models have the LiC score of around $0.6$. 
However, as mentioned in the previous section, this metric actually encourages the model to generate answers each turn prematurely, regardless of the solvability of the current context, which is not desired behavior. 
Therefore, we also introduce the \textit{Abstain Score}, the ratio of the problems that are correctly abstained in the sharded instruction setting, which is calculated as $\frac{Abstention(Unsolvable Sharded)}{Total(Unsolvable Sharded)}$. For this metric, we utilize the LLM-as-a-judge strategy that prompts GPT4o to judge whether the model's answer mentioning the current question is not solvable yet. This metric is more aligned with the goal of trustworthy AI and is more interpretable for humans to understand the model's behavior. Note, although during training, we encourage the model to generate ``\boxed{Abstain}'' for calculating rewards, we avoid utilizing the simple string matching strategy for evaluation for fair comparison with other models. Utilizing the LLM-as-a-judge, we can better evaluate the model's response if it expresses the abstain behavior but does not explicitly use the predefined string.

\subsection{Main Results}

\begin{table*}[t]
    \centering

    \resizebox{\textwidth}{!}{%
    \begin{tabular}{l|cccc|cccc|cccc|cccc|cc}
    \toprule
    \multirow{2}{*}{\textbf{Model}} & \multicolumn{4}{c|}{\textbf{Math}} & \multicolumn{4}{c|}{\textbf{Code}} & \multicolumn{4}{c|}{\textbf{Database}} & \multicolumn{4}{c|}{\textbf{Actions}} & \multicolumn{2}{c}{\textbf{Average}} \\
    \cmidrule(lr){2-5} \cmidrule(lr){6-9} \cmidrule(lr){10-13} \cmidrule(lr){14-17} \cmidrule(lr){18-19}
    & \textbf{Concat} & \textbf{Sharded} & \textbf{LiC} & \textbf{Abstain} & \textbf{C} & \textbf{S} & \textbf{L} & \textbf{A} & \textbf{C} & \textbf{S} & \textbf{L} & \textbf{A} & \textbf{C} & \textbf{S} & \textbf{L} & \textbf{A} & \textbf{LiC} & \textbf{Abstain} \\
    \midrule
    \textbf{OLMo2}~\citep{olmo20242olmo2furious} & 80.1 & 46.3 & 57.8 & - & 16.3 & 14.4 & 88.3 & - & 40.5 & 22.4 & 55.3 & - & 49.8 & 13.8 & 27.7 & - & 57.3 & - \\
    \textbf{Claude3-Haiku}~\citep{anthropic2024claude3ModelCard} & 76.1 & 47.1 & 61.9 & - & 36.3 & 31.5 & 86.8 & - & 76.5 & 31.8 & 41.6 & - & 80.2 & 55.9 & 69.7 & - & 65.0 & - \\
    \textbf{GPT4o-mini}~\citep{openai2024gpt4oMini} & 88.0 & 58.7 & 66.7 & - & 66.7 & 50.3 & 75.4 & - & 90.7 & 40.2 & 44.3 & - & 92.2 & 52.4 & 56.8 & - & 60.8 & - \\
    \textbf{Llama3.3-70B}~\citep{grattafiori2024llama3herdmodels} & 91.8 & 61.5 & 67.0 & - & 52.7 & 51.6 & 97.9 & - & 87.9 & 35.4 & 40.3 & - & 97.0 & 71.0 & 73.2 & - & 69.6 & - \\
    \textbf{Phi-4}~\citep{abdin2024phi} & 90.4 & 52.5 & 58.1 & - & 48.4 & 39.1 & 80.8 & - & 79.6 & 33.1 & 41.6 & - & 76.0 & 34.1 & 44.9 & - & 56.4 & - \\
    \textbf{Llama4-Scout}~\citep{meta2025llama4} & 92.9 & 67.0 & 72.1 & - & 60.3 & 46.4 & 76.9 & - & 81.5 & 27.1 & 33.3 & - & 98.3 & 69.9 & 71.1 & - & 63.4 & - \\
    \textbf{GPT-o3}~\citep{openai2025o3} & 80.0 & 63.1 & 78.9 & - & 87.2 & 53.0 & 60.8 & - & 83.3 & 35.4 & 42.5 & - & 91.5 & 60.2 & 65.8 & - & 62.0 & - \\
    \textbf{Claude3.7-Sonnet}~\citep{anthropic2024claude3ModelCard} & 87.2 & 70.0 & 80.3 & - & 76.2 & 65.6 & 86.1 & - & 81.5 & 34.9 & 42.8 & - & 96.0 & 33.3 & 34.7 & - & 61.0 & - \\
    \textbf{DeepSeek-R1}~\citep{guo2025deepseek} & 92.9 & 67.3 & 72.4 & - & 97.1 & 70.9 & 73.0 & - & 89.9 & 31.5 & 35.0 & - & 97.0 & 47.5 & 49.0 & - & 57.4 & - \\
    \textbf{GPT-4o}~\citep{hurst2024gpt} & 91.9 & 67.9 & 73.9 & 39.9 & 82.9 & 61.3 & 73.9 & 29.8 & 91.7 & 42.3 & 46.1 & 29.4 & 97.1 & 65.0 & 66.9 & 26.8 & 65.2 & 31.5 \\
    \textbf{Gemini-2.5-Flash}~\citep{comanici2025gemini} & 88.4 & 66.1 & 74.8 & 47.0 & 92.5 & 68.3 & 73.8 & 26.8 & 95.5 & 51.3 & 53.7 & 32.8 & 89.2 & 42.6 & 47.8 & 25.4 & 62.5 & 33.0 \\
    \textbf{GPT-4.1}~\citep{openai2025gpt4_1} & 89.7 & 70.7 & 78.8 & 40.5 & 88.7 & 72.6 & 81.8 & 27.4 & 86.5 & 46.0 & 53.2 & 35.8 & 98.5 & 62.9 & 63.9 & 30.1 & 69.4 & 33.5 \\
    \textbf{Gemini-2.5-Pro}~\citep{comanici2025gemini}  & 89.3 & 64.3 & 72.0 & 46.2 & 95.7 & 68.1 & 71.2 & 30.2 & 94.9 & 43.8 & 46.2 & 33.1 & 98.1 & 36.3 & 37.0 & 32.9 & 56.6 & 35.6 \\
    \midrule
    \midrule
    \textbf{Qwen3-1.7B}~\citep{grattafiori2024llama3herdmodels} & 67.6 & 44.4 & 65.7 & 26.8 & 33.4 & 15.0 & 44.9 & 19.8 & 51.1 & 38.8 & 76.0 & 20.4 & 83.1 & 40.2 & 48.4 & 24.9 & 58.8 & 23.0 \\
    \textbf{Qwen3-1.7B (Ours, RLAAR)} & 70.9 & 60.0 & \underline{84.7} & \underline{77.2} & 37.0 & 22.9 & \underline{61.94} & \underline{69.3} & 47.7 & 41.1 & \underline{92.5} & \underline{62.0} & 84.8 & 41.3 & \underline{48.7} & \underline{65.2} & \underline{71.9} & \underline{68.4} \\
    \midrule
    \textbf{Qwen3-8B}~\citep{grattafiori2024llama3herdmodels} & 88.4 & 72.8 & 82.4 & 39.1 & 60.0 & 35.0 & 58.3 & 28.9 & 76.0 & 46.1 & 60.6 & 31.1 & 99.0 & 48.5 & 49.0 & 34.7 & 62.6 & 33.5 \\
    \textbf{Qwen3-8B} (SFT, Distill Gemini-2.5-Pro) & 87.4 & 75.6 & 86.5 & 57.4 & 58.4 & 34.2 & 58.6 & 42.9 & 67.6 & 40.2 & 59.5 & 47.8 & 90.2 & 41.6 & 46.1 & 47.8 & 62.7 & 49.0 \\
    \textbf{Qwen3-8B (Ours, RLAAR)} & 87.2 & 82.2 & \underline{94.3} & \underline{80.8} & 63.2 & 44.8 & \underline{70.9} & \underline{72.3} & 77.6 & 63.2 & \underline{81.4} & \underline{69.7} & 97.4 & 52.3 & \underline{53.7} & \underline{69.7} & \underline{75.1} & \underline{\textbf{73.4}} \\
    \midrule
    \textbf{Qwen2.5-7B}\citep{qwen2025qwen25technicalreport} & 82.2 & 59.8 & 72.8 & 36.1 & 55.0 & 31.0 & 56.4 & 26.2 & 73.2 & 38.5 & 52.5 & 27.7 & 95.8 & 35.0 & 36.5 & 31.9 & 54.6 & 30.5 \\
    \textbf{Qwen2.5-7B} (SFT, Distill Gemini-2.5-Pro) & 82.7 & 60.3 & 72.9 & 57.2 & 56.4 & 33.9 & 60.1 & 47.7 & 69.6 & 36.4 & 52.3 & 42.9 & 91.7 & 31.6 & 34.5 & 52.7 & 55.0 & 50.1 \\
    \textbf{Qwen2.5-7B (Ours, RLAAR)} & 83.4 & 79.2 & \underline{95.0} & \underline{77.0} & 58.7 & 41.2 & \underline{70.2} & \underline{72.0} & 74.9 & 65.4 & \underline{87.3} & \underline{61.3} & 92.8 & 49.3 & \underline{53.0} & \underline{70.3} & \underline{\textbf{76.4}} & \underline{67.6} \\
    \bottomrule
    \end{tabular}%
    }
    \vspace{-2.2mm}
    \caption{Performance comparison of previous well-known models. From OLMo2 to Gemini-2.5-Pro, we directly borrow the results from the original LiC paper \citep{laban2025llms}, except for the abstain scores. \textit{Concat} and \textit{Sharded} represent the performance of the concatenated instruction setting and the sharded instruction setting, respectively; \textit{LiC} and \textit{Abstain} represent the LiC score and the abstain score, respectively. The better performances compared with the baseline models are underlined, and the best average performances across all models are highlighted in bold.}
    \vspace{-1em}
    \label{tbl:main}
    \end{table*}

Table \ref{tbl:main} shows the performance comparison of previous well-known models. Performance on $4$ different tasks is presented, including math, code, database, and actions; \textit{Concat} and \textit{Sharded} represent the performance of the concatenated instruction setting and the sharded instruction setting, respectively; \textit{LiC} and \textit{Abstain} represent the LiC score and the Abstain score, respectively. The better performance compared with the baseline models is underlined, and the best performance across all models is highlighted in bold.

As shown in the table, even the recent state-of-the-art models like GPT-4.1 and Gemini-2.5-Pro have the average LiC score of around $0.6$, which means their performance can only maintain $60\%$ of the performance of the concatenated instruction setting, which shows the severity of the ``Lost in Conversation'' problem. 
Moreover, the performance comparison shows that supervised finetuning on relatively strong models like Gemini-2.5-Pro is not effective for mitigating the LiC problem.
On the other hand, models trained with our method outperform the baseline models on all tasks to a large margin, represented by the underlined results. Even when compared with the well-known state-of-the-art models, our method can reach the average LiC score of around $0.75$, and for math tasks, we can even reach the LiC score of around $0.9$, which shows the effectiveness of our method, highlighted by the bold results.

Moreover, our method not only focuses on the LiC score, but also tries to improve the capability of the models to justify the solvability of the current context, thus bringing improvement to the trustworthy AI, which is reflected by the improvement of the abstain score. While most of the models only achieve the abstain score of around $0.2 - 0.4$, indicating that they tend to answer questions regardless of the solvability, our method can achieve the abstain score of around $0.7 - 0.8$, indicating that the model has the capability to justify the solvability of the current context. For more qualitative details, failure and success examples are provided in Appendix \ref{sec:examples}.

\subsection{Ablation Studies}

\subsubsection{Effect of the Abstention Reward}

\begin{table}[!t]
    \centering
    \resizebox{0.48\textwidth}{!}{%
    \begin{tabular}{l|cc}
    \toprule
    \multicolumn{1}{l|}{\textbf{Abstain Ratio}} & \textbf{LiC score} & \textbf{Abstain score}  \\
    \midrule
    Qwen3-1.7B (Baseline) & 58.8 & 23.0\\
    \midrule
    Qwen3-1.7B ($m=0$, No Abstention) & 64.3 & 17.4 \\
    Qwen3-1.7B ($m=0.1$, Default) & \textbf{71.9} & 68.4 \\
    Qwen3-1.7B ($m=0.2$) & 70.8 & 69.5 \\
    Qwen3-1.7B ($m=0.3$) & 62.6 & 72.8 \\
    Qwen3-1.7B ($m=0.4$) & 47.1 & 80.3 \\
    Qwen3-1.7B ($m=0.5$) & 27.4 & \textbf{92.8} \\
    \bottomrule
    \end{tabular}
    }
    \vspace{-2.2mm}
    \caption{Ablation studies on the effect of the abstention reward. The baseline model is Qwen3-1.7B, and the abstention ratio $m$ is varied from $0$ to $0.5$, all other settings are the same as the default setting. The abstention ratio $m$ represents the ratio of the rollouts in each step that are categorized as~\textit{Unsolvable-Multi}, thus using the abstention reward rather than the accuracy reward. The best performances are highlighted in bold.}
    \vspace{-1em}
    \label{tbl:abs}
    
\end{table}

In this section, we conduct ablation studies to investigate the effect of the abstention reward on the performance of the model. As shown in Table \ref{tbl:abs},  the baseline model is Qwen3-1.7B, and the abstention ratio $m$ is varied from $0$ to $0.5$; all other settings are the same as the default setting. The abstention ratio $m$ represents the ratio of the rollouts in each step that are categorized as~\textit{Unsolvable-Multi}, thus using the abstention reward rather than the accuracy reward. \looseness-1

When $m=0$, the model is only trained on the \textit{Solvable-Multi} rollouts using the accuracy reward. Compared with the baseline model, when $m=0.5$, the model can already achieve the abstain score of $64.3\%$, which is benefited from the explicit usage of multi-turn rollouts, which forces the model to get the necessary information across previous multi-turn context to solve the problem. However, in this strategy, the model tends to try answering the question in every turn, which causes the growth and potential pollution of the previous context. Thus, the performance is not as good as the default setting. \looseness-1

Thus, when $m>0$, the models are trained on both accuracy reward and abstention reward, which represents our main method, RLAAR. 
Through the performance comparison from $m=0.1$ to $m=0.5$, we can further observe that the LiC score is gradually decreased while the abstain score is gradually increased. It happens because the abstain strategy is much easier for LLMs to learn; thus, when the ratio of the \textit{Unsolvable-Multi} rollouts is too large, the model tends to overfit the abstain strategy and thus abstain from every request without really solving the problem. In this case, the high abstention score is not useful. 

\subsubsection{Effect of the Curriculum Learning}

\begin{table}[!t]
    \centering
    \resizebox{0.48\textwidth}{!}{%
    \begin{tabular}{l|cc|c}
    \toprule
    \multicolumn{1}{l|}{\textbf{Curriculum Setting}} & \textbf{LiC score} & \textbf{Abstain score} & \textbf{Steps}  \\
    \midrule
    Qwen3-1.7B (Baseline) & 58.8 & 23.0 \\
    \midrule
    Qwen3-1.7B (No Curriculum) & 63.2 & 57.8 & 0 \\
    Qwen3-1.7B ($\rho=1.0$) & 65.6 & 64.3 & 1000+ \\
    Qwen3-1.7B ($\rho=0.9$) & \textbf{72.2} & 66.4 & 470 \\
    Qwen3-1.7B ($\rho=0.8$, Default) & 71.9 & \textbf{68.4} & 90 \\
    Qwen3-1.7B ($\rho=0.7$) & 62.1 & 60.7 & 70 \\
    Qwen3-1.7B ($\rho=0.6$) & 40.9 & 47.6 & 40 \\
    \bottomrule
    \end{tabular}
    }
    \vspace{-2.2mm}
    \caption{Ablation studies on the effect of the curriculum learning strategy. The baseline model is Qwen3-1.7B, and the threshold ratio $\rho$ is varied from $0$ to $0.5$, all other settings are the same as the default setting. The threshold ratio $\rho$ represents the ratio of the baseline threshold that the model's reward should exceed to progress to the next difficulty level.
    The best performances are highlighted in bold. In addition to the performance, we also report the steps required to reach the Randomized Training stage (Stage 3), to provide a better understanding of the curriculum learning strategy.
    }
    \vspace{-1em}
    \label{tbl:cl}
    
\end{table}



In this section, we conduct ablation studies to investigate the effect of the curriculum learning strategy. As shown in Table \ref{tbl:cl}, the baseline model is Qwen3-1.7B, and the threshold ratio $\rho$ is varied from $0$ to $0.5$; all other settings are the same as the default setting. The threshold ratio $\rho$ represents the ratio of the baseline threshold that the model's reward should exceed to progress to the next difficulty level. A higher $\rho$ means the model needs to reach a higher average reward to progress to the next difficulty level. 

When no curriculum is applied, the model is trained directly as in Randomized Training stage (Stage 3), where a random number of shards $K$ is sampled between 1 and $K_{\text{max}}$ for each step. As shown in the table, the model has already reached promising performance on both LiC score and abstain score. However, the training process is not as stable as when using the curriculum learning strategy, as it is hard for the model to get positive feedback in the early stage. On the contrary, curriculum learning is naturally suitable for this shard instruction setting, as we can gradually increase the number of shards/turns for the model to train on. However, when $\rho=1$, which means the model needs to reach an average reward equal to training on a single-turn setting to progress to the next difficulty level, the model gets stuck in every difficulty level, as it's very hard for the model to reach the single-turn setting performance. As a result, the model needs more than $1000$ steps to overfit to the training data and thus pass all the difficulty levels. When $\rho$ is lower than $1$, it means the model only needs to reach the average reward of $\rho$ times the baseline threshold to progress to the next difficulty level, which is easier. As a result, the model can progress to the next difficulty level in a more stable way. Comparing $\rho=0.9$ and $\rho=0.8$, we can observe that the model performance is similar, but the model needs many fewer steps to pass all the difficulty levels when $\rho=0.8$, thus we set $\rho=0.8$ as the default setting. On the other hand, when $\rho$ is too small, the model may pass all the difficulty levels too fast to really learn things.








\section{Conclusion}

We introduced Curriculum Reinforcement Learning with Verifiable Accuracy and Abstention Rewards (RLAAR) to mitigate the Lost-in-Conversation (LiC) phenomenon in multi-turn language model interactions. By combining verifiable accuracy and abstention rewards within a competence-gated curriculum, our framework enables models to learn when to act and when to defer, leading to more stable and trustworthy dialogue behavior. Experiments demonstrate that RLAAR consistently reduces LiC degradation, enhances reliability through calibrated abstention.

\clearpage
\section*{Limitation}

While our method demonstrates significant improvements in mitigating Lost-in-Conversation (LiC), our experiments are conducted primarily on math and code datasets, which may not fully capture the complexity of open-domain or knowledge-intensive dialogues. Extending RLAAR to broader conversational settings, such as factual QA or multi-agent collaboration, remains a future direction.

\bibliography{custom}

\appendix

\clearpage
\section{Discussion on Evaluation Metrics}

For the Abstain Score, we use GPT-4o as an LLM judge. Importantly, the judge is not asked to assess subjective response quality; it only determines whether the response expresses \emph{abstention intent}, e.g., that the question cannot be answered from the available context or that more information is needed. In this sense, LLM-as-a-judge serves as a semantic classifier for abstention behavior.

This also explains why training and evaluation use different criteria. During training, we follow standard RLVR and use deterministic string matching, treating \texttt{Abstain} as a verifiable terminal action. For evaluation, however, exact matching would unfairly penalize baseline models that were not trained to emit the literal token \texttt{Abstain} and may instead express uncertainty in natural language. Thus, LLM-as-a-judge is used only to normalize semantically equivalent abstention expressions across models. 

We further validate this choice with human evaluation on 100 math examples (50 solvable and 50 unsolvable), comparing human labels with GPT-4o judgments. As shown in Table~\ref{tbl:judge}, the agreement is high for both the Qwen2.5-7B baseline (0.92) and our Qwen2.5-7B RLAAR model (0.97), supporting the reliability of this metric.

\begin{table}[h]
    \centering
    \resizebox{0.42\textwidth}{!}{%
    \begin{tabular}{l|c}
    \toprule
    \textbf{Model} & \textbf{Human--LLM Consistency} \\
    \midrule
    Qwen2.5-7B Baseline & 0.92 \\
    Qwen2.5-7B (Ours, RLAAR) & 0.97 \\
    \bottomrule
    \end{tabular}
    }
    \vspace{-2.2mm}
    \caption{Agreement between human annotations and GPT-4o on whether a response expresses abstention. We randomly sample 100 math examples in total, including 50 solvable and 50 unsolvable cases.}
    \vspace{-1em}
    \label{tbl:judge}
\end{table}

\section{Examples}
\label{sec:examples}

In this section, we provide a failure example of the baseline model and a success example of our method to show the desired behavior of the model.

Example \ref{box:conversation_failure} shows a typical failure example of most of the existing models. Every time the user provides some new information, the model tends to answer the question prematurely without checking the solvability of the question, and thus results in an incorrect answer.

On the other hand, Example \ref{box:conversation_success} shows a success example of our method, which shows the desired behavior when facing this kind of situation. Every time the user provides some new information, the model will check the solvability of the question first. If the question is not solvable, the model will abstain from the question. Only if the model is sure that the question is solvable, it answers the question.

\onecolumn

\begin{tcolorbox}[title=Failure Example, label={box:conversation_failure}]
    \textit{\textbf{(Round 1) User: how many vines does Steve need to grow his cherry tomatoes? }}\\
    \textbf{Assistant:} To determine how many vines Steve needs to grow his cherry tomatoes, we need to know: 1. How many cherry tomato plants Steve wants to grow. 2. How many vines each cherry tomato plant requires. Step 1: Assume a standard setup - Each cherry tomato plant typically requires 1 vine to grow. - Each vine can produce approximately 3-4 cherry tomato plants. Step 2: Estimate based on the number of plants Let’s assume Steve wants to grow 10 cherry tomato plants. - If each plant needs 1 vine, then Steve needs 10 vines. Step 3: Final Answer 10 So, Steve needs 10 vines to grow his cherry tomatoes.\\
    \textcolor{red}{(Premature Answer)}

    \vspace{0.3em}

    \textit{\textbf{(Round 2) User: He eats 6 tomatoes every day.} }\\
    \textbf{Assistant:} Let's determine how many tomatoes Steve eats in a week. Step 1: Calculate daily consumption - Steve eats 6 tomatoes per day. Step 2: Calculate weekly consumption - There are 7 days in a week. - So, total tomatoes eaten in a week = 6 \text{ tomatoes/day} $\times 7 \text{ days}$. $6 \times 7 = 42$. Step 3: Final Answer 42 Steve eats 42 tomatoes in a week.\\
    \textcolor{red}{(Premature Answer)}
    \vspace{0.3em}

    \textit{\textbf{(Round 3) User: Steve wants to grow his own cherry tomatoes to eat more.} }\\
    \textbf{Assistant:} To determine how many vines Steve needs to grow his own cherry tomatoes, we need to make some assumptions and use logical reasoning. Step 1: Assume the number of plants Steve wants to grow Let’s assume Steve wants to grow 10 cherry tomato plants. Step 2: Determine how many vines are needed per plant Each cherry tomato plant typically requires 1 vine to grow. Step 3: Total number of vines needed $\text{Number of vines} = \text{Number of plants} \times \text{Vines per plant} = 10 \times 1 = 10$ Step 4: Consider the tomato yield per vine Each vine can produce 3–4 cherry tomato plants, so: $\text{Tomato yield per vine} = 3–4 \text{ plants}$ If Steve wants to grow 10 plants, and each vine can yield 3 plants, then: $\text{Number of vines} = \frac{10}{3} \approx 3.33$ Since you can’t grow a fraction of a vine, Steve would need 4 vines to grow 12 cherry tomato plants. Final Answer: 4 Steve needs 4 vines to grow his own cherry tomatoes.\\
    \textcolor{red}{(Premature Answer)}
    \vspace{0.3em}

    ...
    \vspace{0.3em}

    \textit{\textbf{(Round Final) User: Each vine can produce 3 tomatoes weekly. }}\\
    \textbf{Assistant:} To determine how many vines Steve needs to produce 30 tomatoes per week, we follow these steps: Step 1: Understand the production rate - Each vine produces 3 tomatoes per week. Step 2: Calculate the number of vines needed $\text{Number of vines} = \frac{\text{Total tomatoes needed}}{\text{Tomatoes per vine}} = \frac{30}{3} = 10$ Final Answer: 10 Steve needs 10 vines to produce 30 tomatoes per week.\\
    \textcolor{red}{(Incorrect Answer)}

    \end{tcolorbox}

\begin{tcolorbox}[title=Success Example, label={box:conversation_success}]
        \textit{\textbf{(Round 1) User: how many vines does Steve need to grow his cherry tomatoes? }}\\
        \textbf{Assistant:} To determine how many vines Steve needs to grow his cherry tomatoes, we would need more specific information, such as: - The number of cherry tomato plants Steve wants to grow. - The number of vines each plant requires. - The number of plants per vine (if applicable). - Any constraints or limitations (e.g., space, resources, etc.). Since none of these details are provided in the question, it is not possible to determine the exact number of vines Steve needs. Reasoning: - Without knowing how many plants Steve wants, or how many vines each plant requires, we cannot compute the number of vines. - Similarly, if there are multiple plants per vine, this would also affect the total number of vines needed. Conclusion: The question is not solvable with the information given. \boxed{Abstain}

        \textcolor{red}{(Abstain)}
    
        \vspace{0.3em}
    
        \textit{\textbf{(Round 2) User: He eats 6 tomatoes every day.} }\\
        \textbf{Assistant:} We are given the following information: - Steve eats 6 tomatoes every day. Step 1: Understand the problem We are asked to find how many vines Steve needs to grow his cherry tomatoes. However, we do not have any information about: - How many plants Steve wants to grow. - How many tomatoes each plant produces. - How many vines are needed per plant. Step 2: Identify missing information The question is not solvable unless we have the following: - The number of cherry tomato plants Steve wants to grow. - The number of tomatoes each plant produces. - The number of vines required per plant. Step 3: Determine if the question is solvable Since we do not have the above information, we cannot compute how many vines Steve needs. Conclusion: The question cannot be solved without additional information. \boxed{Abstain} \\
        \textcolor{red}{(Abstain)}
        \vspace{0.3em}
    
        \textit{\textbf{(Round 3) User: Steve wants to grow his own cherry tomatoes to eat more.} }\\
        \textbf{Assistant:} To determine how many vines Steve needs to grow his own cherry tomatoes, we must first understand the relationship between vines, tomatoes, and Steve's needs. Step 1: Understand the context Steve is planning to grow his own cherry tomatoes to supply his needs. This implies: - He wants to grow cherry tomatoes. - He does not need to buy tomatoes from a store. - He is planting and growing his own tomatoes. Step 2: Gather relevant information The key variables we need to know are: - How many tomatoes Steve needs per day: From the previous question, we know that Steve eats 6 tomatoes every day. - How many tomatoes each vine produces: This information is not provided in the question. - How many vines are needed per plant: This information is not provided in the question. Step 3: Evaluate the problem Without knowing: - How many tomatoes Steve needs in total (i.e., total daily or monthly consumption), - How many tomatoes each vine produces, - How many vines are required per plant, we cannot compute the exact number of vines Steve needs. Step 4: Determine if the problem is solvable The question is not solvable due to a lack of necessary data. It is a mathematical problem that requires unknown parameters to compute the answer. Final Conclusion: The question cannot be solved with the information provided. \boxed{Abstain}
        \\
        \textcolor{red}{(Abstain)}
        \vspace{0.3em}
    
        ...
        \vspace{0.3em}
    
        \textit{\textbf{(Round Final) User: Each vine can produce 3 tomatoes weekly. }}\\
        \textbf{Assistant:} To determine how many vines Steve needs to grow his cherry tomatoes, we now use the following known information: Step 1: Known Information - Steve eats 6 tomatoes every day. - Steve eats twice as many tomatoes as his girlfriend, so: $\text{Girlfriend's daily consumption} = \frac{6}{2} = 3 \text{ tomatoes/day}$ $\text{Total daily consumption} = 6 + 3 = 9 \text{ tomatoes/day}$ - Each vine can produce 3 tomatoes every week. Step 2: Convert Weekly Yield to Daily Yield Since each vine produces 3 tomatoes per week, we calculate the daily yield per vine: $\text{Daily yield per vine} = \frac{3 \text{ tomatoes/week}}{7 \text{ days/week}} = \frac{3}{7} \text{ tomatoes/day}$ Step 3: Calculate Total Daily Tomatoes Needed We have already determined: $\text{Total daily tomatoes needed} = 9 \text{ tomatoes/day}$ Step 4: Calculate Total Number of Vines Needed Now, we calculate the number of vines required to produce the 9 tomatoes per day. $\text{Number of vines needed} = \frac{9}{\frac{3}{7}} = 9 \times \frac{7}{3} = 21 \text{ vines}$ Final Answer: \boxed{21}\\
        \textcolor{red}{(Correct Answer)}
    
        \end{tcolorbox}

\twocolumn

\section{Pseudo Code}
\label{sec:pseudocode}

The pseudo code of the RLAAR method is shown in Algorithm \ref{alg:rlaar} and Algorithm \ref{alg:rollouts}.

\begin{algorithm*}[t]
    \caption{Curriculum RL with Verifiable Accuracy and Abstention Rewards (RLAAR)}
    \label{alg:rlaar}
    \KwIn{Policy $\pi_{\theta}$; Dataset $\mathcal{D}$; Max shards $K_{\max}$; 
    Abstention ratio $m$; Threshold ratio $\rho$; Window size $w$; Batch size $B$}
    \KwOut{Trained policy $\pi_{\theta}$}
    
    \BlankLine
    \textbf{Stage 1: Threshold Establishment}\;
    $K \leftarrow 1$\;
    \For{$i=1$ to $w$}{
        trajectories $\leftarrow [\,]$, rewards $\leftarrow [\,]$\;
        \For{$j = 1$ to $B$}{
            $\text{task} \sim \text{Sample}(\mathcal{D})$\;
            $\tau \leftarrow \text{SolvableSingleRollout}(\pi_{\theta}, \text{task}, K{=}1)$\;
            $r \leftarrow r_{\text{acc}}(\tau)$\;
            trajectories.append($\tau$); rewards.append($r$)\;
        }
        $\pi_{\theta} \leftarrow \text{UpdatePolicy}(\pi_{\theta}, \text{trajectories}, \text{rewards})$\;
    }
    $\bar{r}_0 \leftarrow \text{MovingAverage}(\text{all\_rewards}, \text{window}=w)$, $\text{threshold} \leftarrow \rho \times \bar{r}_0$\;
    
    \BlankLine
    \textbf{Stage 2: Main Training (Curriculum)}\;
    $K \leftarrow 2$\;
    \While{$K \leq K_{\max}$}{
        trajectories $\leftarrow [\,]$, rewards $\leftarrow [\,]$\;
        \For{$i = 1$ to $B$}{
            $\text{task} \sim \text{Sample}(\mathcal{D})$\;
            \If{$\text{random()} < m$}{
                $\tau \leftarrow \text{UnsolvableMultiRollout}(\pi_{\theta}, \text{task}, M)$\;
                $r \leftarrow r_{\text{abs}}(\tau)$\;
            }
            \Else{
                $\tau \leftarrow \text{SolvableMultiRollout}(\pi_{\theta}, \text{task}, K)$\;
                $r \leftarrow r_{\text{acc}}(\tau)$\;
            }
            trajectories.append($\tau$); rewards.append($r$)\;
        }
        $\pi_{\theta} \leftarrow \text{UpdatePolicy}(\pi_{\theta}, \text{trajectories}, \text{rewards})$,
        $\bar{r} \leftarrow \text{MovingAverage}(\text{rewards}, \text{window}=w)$\;
        \If{$\bar{r} \geq \text{threshold}$}{
            $K \leftarrow K + 1$\;
        }
    }
    
    \BlankLine
    \textbf{Stage 3: Randomized Training}\;
    \For{$i=1$ to Max\_steps}{
        $K \leftarrow \text{random\_int}(1, K_{\max})$, trajectories $\leftarrow [\,]$, rewards $\leftarrow [\,]$\;
        \For{$i = 1$ to $B$}{
            $\text{task} \sim \text{Sample}(\mathcal{D})$\;
            \If{$\text{random()} < m$}{
                $\tau \leftarrow \text{UnsolvableMultiRollout}(\pi_{\theta}, \text{task}, M)$\;
                $r \leftarrow r_{\text{abs}}(\tau)$\;
            }
            \Else{
                $\tau \leftarrow \text{SolvableMultiRollout}(\pi_{\theta}, \text{task}, K)$\;
                $r \leftarrow r_{\text{acc}}(\tau)$\;
            }
            trajectories.append($\tau$); rewards.append($r$)\;
        }
        $\pi_{\theta} \leftarrow \text{UpdatePolicy}(\pi_{\theta}, \text{trajectories}, \text{rewards})$\;
    }
    \Return{$\pi_{\theta}$}\;
    \end{algorithm*}

\begin{algorithm*}[t]
        \caption{Multi-turn Rollout Functions}
        \label{alg:rollouts}
        \DontPrintSemicolon
        \SetKwFunction{FSingle}{SolvableSingleRollout}
        \SetKwFunction{FMulti}{SolvableMultiRollout}
        \SetKwFunction{FUnsolv}{UnsolvableMultiRollout}
        \SetKwProg{Fn}{Function}{:}{end}
        
        \Fn{\FSingle{$\pi_{\theta}$, task $T$, $K{=}1$}}{
          $s_{1} \leftarrow \textsc{FullQuestion}(T)$\;
          $a_{1} \sim \pi_{\theta}(\cdot \mid s_{1})$\;
          $\tau \leftarrow [\, (s_{1}, a_{1}) \,]$\;
          \Return{$\tau$}\;
        }
        
        \BlankLine
        \Fn{\FMulti{$\pi_{\theta}$, task $T$, $K$}}{
          $S \leftarrow \{s_{1}, s_{2}, \dots, s_{K}\} \leftarrow \textsc{ShardQuestion}(T, K)$\;
          $\tau \leftarrow [\,]$;\quad $c_{1} \leftarrow s_{1}$\;
          \For{$k \leftarrow 1$ \KwTo $K$}{
            $a_{k} \sim \pi_{\theta}(\cdot \mid c_{k})$\;
            $\tau.\texttt{append}\big((c_{k}, a_{k})\big)$\;
            \If{$k < K$}{
              $c_{k+1} \leftarrow (s_{1}, a_{1}, \dots, s_{k}, a_{k}, s_{k+1})$\;
            }
          }
          \Return{$\tau$}\;
        }
        
        \BlankLine
        \Fn{\FUnsolv{$\pi_{\theta}$, task $T$, $M$}}{
          $S \leftarrow \{s_{1}, s_{2}, \dots, s_{M}\} \leftarrow \textsc{ShardQuestion}(T, M, \textit{incomplete}{=}\texttt{True})$\;
          $\tau \leftarrow [\,]$;\quad $c_{1} \leftarrow s_{1}$\;
          \For{$k \leftarrow 1$ \KwTo $M$}{
            $a_{k} \sim \pi_{\theta}(\cdot \mid c_{k})$\;
            $\tau.\texttt{append}\big((c_{k}, a_{k})\big)$\;
            \If{$k < M$}{
              $c_{k+1} \leftarrow (s_{1}, a_{1}, \dots, s_{k}, a_{k}, s_{k+1})$\;
            }
          }
          \Return{$\tau$}\;
        }
        \end{algorithm*}

\end{document}